\documentclass[letterpaper, 10 pt, conference]{ieeeconf}

\def\ps@IEEEtitlepagestyle{
  \def\@oddfoot{\mycopyrightnotice}
  \def\@evenfoot{}
}
\def\mycopyrightnotice{
  {\footnotesize IEEE ICRA-2022, May.23-27. Philadelphia (PA), USA.~\copyright~IEEE All rights reserved. \hfill} 
  \gdef\mycopyrightnotice{}
}

\usepackage{eso-pic}
\newcommand\AtPageUpperMyleft[1]{\AtPageUpperLeft{
 \put(\LenToUnit{0.15\paperwidth},\LenToUnit{-1.5cm}){
     \parbox{0.9\textwidth}{\raggedleft\fontsize{10}{11}\selectfont #1}}
 }}
\newcommand{\conf}[1]{
\AddToShipoutPictureBG*{
\AtPageUpperMyleft{#1}
}
}
 
\conf{\textit{Accepted Version}: 2022 IEEE International Conference on Robotics and Automation (ICRA), May.23-27, 2022 - Philadelphia (PA), USA.
~\textcopyright 2022 IEEE. Personal use of this material is permitted. Permission from IEEE must be obtained for all other uses. More information \href{http://www.ieee.org/publications_standards/publications/rights/index.html}{\textit{here}}}
\newcommand{\placetextbox}[3]{
\setbox0=\hbox{#3}
\AddToShipoutPictureFG{ \put(\LenToUnit{#1\paperwidth},\LenToUnit{#2\paperheight}){\vtop{{\null}\makebox[0pt][c]{#3}}}}
}
\placetextbox{.372}{0.047}{~\copyright~IEEE All rights reserved. IEEE-ICRA-2022, May.23-27. Philadelphia (PA), USA.\hfill}

\IEEEoverridecommandlockouts
\usepackage{epsfig} 
\graphicspath{{Figures/}}
\usepackage{amsmath} 
\usepackage{amssymb}  
\usepackage[pdfusetitle]{hyperref}
\usepackage{subfigure}
\usepackage{multirow}
\usepackage{makecell}
\usepackage[colorinlistoftodos]{todonotes}
\usepackage{xcolor}
\hypersetup{
	colorlinks=true,
	linkcolor=black,
	citecolor=black,
}
\usepackage{soul}
\soulregister\cite7 
\soulregister\citep7 
\soulregister\citet7 
\soulregister\ref7 
\soulregister\pageref7 

\usepackage[separate-uncertainty=true,multi-part-units=single]{siunitx}
\usepackage{booktabs}
\usepackage{environ}
\usepackage{tikz}
\usetikzlibrary{matrix, chains, positioning, fit, arrows, arrows.meta, shapes, calc}
\usetikzlibrary{decorations.text, decorations.pathmorphing, decorations.pathreplacing, external}

\makeatletter
\newsavebox{\measure@tikzpicture}
\NewEnviron{scaletikzpicturetowidth}[1]{%
  \def\tikz@width{#1}%
  \begin{lrbox}{\measure@tikzpicture}%
  \BODY
  \end{lrbox}%
  \pgfmathparse{#1/\wd\measure@tikzpicture}%
  \BODY
}
\makeatother

\usepackage{array}
\newcommand{\PreserveBackslash}[1]{\let\temp=\\#1\let\\=\temp}
\newcolumntype{C}[1]{>{\PreserveBackslash\centering}m{#1}}
\newcolumntype{R}[1]{>{\PreserveBackslash\raggedleft}m{#1}}
\newcolumntype{L}[1]{>{\PreserveBackslash\raggedright}m{#1}}

\newcount\tmpnum

\def\storedataA#1{\advance\tmpnum by1
   \ifx\end#1\else
      \expandafter\def\csname data:\tmp:\the\tmpnum\endcsname{#1}%
      \expandafter\storedataA\fi
}
\def\getdata[#1]#2{\csname data:\string#2:#1\endcsname}

\title{Unfreezing Social Navigation: Dynamical Systems based Compliance for Contact Control in Robot Navigation}

\author{Diego {Paez-Granados}$^{1\dagger}$, Vaibhav Gupta $^{2}$, Aude Billard$^{2}$
\thanks{$^{\dagger}$ is the corresponding author.
 $^1$ D.Paez-Granados was with LASA, EPFL during this work, and currently is with the SCAI Lab at SPZ, ETH Zurich, Switzerland
    {\tt\small dfpg@ieee.org}}
\thanks{V. Gupta, and A. Billard are with the Learning Algorithms and Systems Laboratory (LASA), Swiss Federal School of Technology in Lausanne -
EPFL, Switzerland
        {\tt\small\{vaibhav.gupta; aude.billard\}@epfl.ch}}
 	}

\begin{document}

\maketitle
\thispagestyle{empty}
\pagestyle{empty}
\addtolength{\textfloatsep}{-0.22in}
\addtolength{\abovecaptionskip}{-0.11in}
\begin{abstract}
Large efforts have focused on ensuring that the controllers for mobile service robots follow proxemics and other social rules to ensure both safe and socially acceptable distance to pedestrians.
Nonetheless, involuntary contact may be unavoidable when the robot travels in crowded areas or when encountering adversarial pedestrians. Freezing the robot in response to contact might be detrimental to bystanders' safety and prevents it from achieving its task. Unavoidable contacts must hence be controlled to ensure the safe and smooth travelling of robots in pedestrian alleys.
We present a force-limited and obstacle avoidance controller integrated into a time-invariant dynamical system (DS) in a closed-loop force controller that let the robot react instantaneously to contact or to the sudden appearance of pedestrians. Mitigating the risk of collision is done by modulating the velocity commands upon detecting a contact and by absorbing part of the contact force through active compliant control when the robot bumps inadvertently against a pedestrian.
We evaluated our method with a personal mobility robot -Qolo- showing contact mitigation with passive and active compliance. We showed the robot able to overcome an adversarial pedestrian within 9 N of the set limit contact force for speeds under 1 m/s.
Moreover, we evaluated integrated obstacle avoidance proving the ability to advance without incurring any other collision.
\end{abstract}



%
\IEEEpeerreviewmaketitle

\section{Introduction}
As a consequence of the widespread diffusion of mobile robots in public environments, the chances that robots and bystanders may collide will increase. Looking at the statistics in the automotive sector, the number of crashes involving pedestrians is a considerable portion of the total number of accidents \cite{Anderson2016}. 
Real-life implementations of any obstacle avoidance are limited by the reactivity of the robot i.e. kinematic and dynamic constraints, actuation power, and highly limited computational resources on mobile robots.
Robots cannot behave like pedestrians. Most of the service robots are non-holonomic and cannot step on the side. They are also less reactive and lacks the means to communicate their intended move in easily interpretable ways, and are rarely knowledgeable of proxemics and other social rules \cite{Charalampous2017}. 
Nonetheless, the utility of mobile service robots is getting traction and valuable services such as in-hospital assistance, last-mile deliveries (Starship Inc. USA), autonomous cleaning robots (Bluebotics, Switzerland) and autonomous wheelchairs (Whill Inc. Japan) are becoming popular.

Similarly to what happened with industrial collaborative robots \cite{Rosenstrauch2017, Haddadin2017}, chances are high that deployment will proceed and that society will slowly accept and allow physical contact between mobile robots and humans, especially in crowded environments \cite{Shrestha2015}.
Hence, there is some urgency to develop control approaches and design requirements to mitigate risks and allow motion control in post-contact between a pedestrian and a service robot.

\begin{figure}[t]
    \centering
    \includegraphics[width=8.5cm]{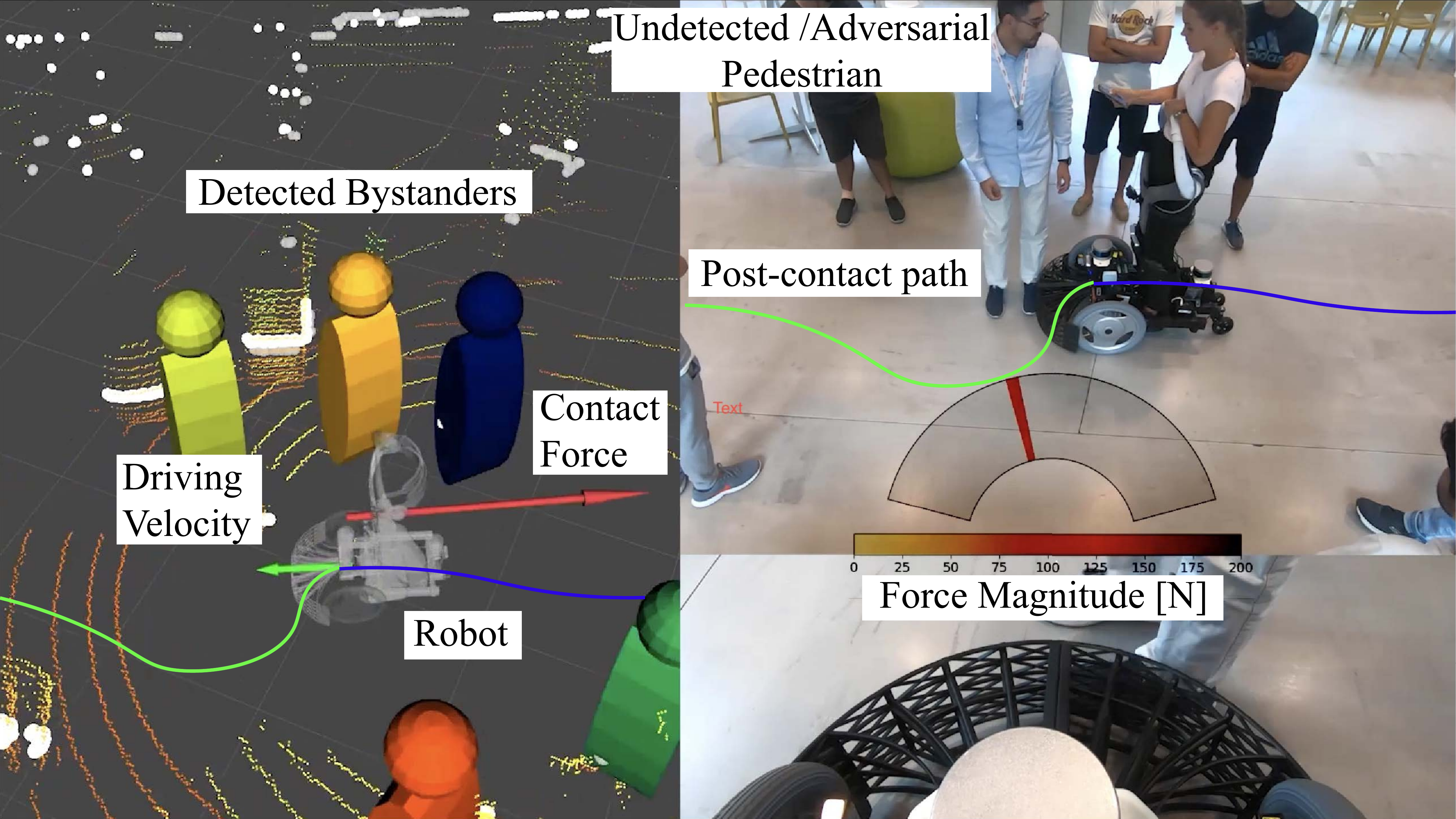}
    \caption{An adversarial pedestrian colliding with a service robot that senses the collision and absorbs part it of through active compliance. The robot finds a feasible path by sliding with a contact-aware surface towards the opposite direction of detected pedestrians. }\label{fig:intro}
\end{figure}

Most attention has been given to pre-collision planning \cite{Haddadin2017,Bajcsy2021} and human motion prediction \cite{Kobayashi2021, Fridovich2020}.
As far as physical collisions between humans and robots are concerned, efforts were solely directed at evaluating collisions with collaborative robot arm manipulators \cite{Haddadin2008, Rosenstrauch2017, Haddadin2009-IJRR}. These works were instrumental and led to the ISO 15066:2016 \cite{ISO15066} standard that establishes values for pain thresholds for blunt impacts with respect to force and pressure, after collisions with a robotic arm for determining what should be its operational velocity around humans. However, there is no equivalent standard for collision with light-weight vehicles \cite{Salvini2021-THRI}.

Mutual anticipation remains critical in ensuring that humans can navigate safely in crowded environments \cite{Murakami2021}. Nonetheless, pedestrians' expectations of robots' motion capabilities may exceed robots' actual motion constraints, thereby increasing the risk of misinterpreted reciprocal avoidance.
Currently, the basic safety control system let the robot freeze as soon as it perceives a contact \cite{Sathyamoorthy2020}, a reaction that would most likely be unexpected by pedestrians and lead to more dangerous collisions with pedestrians stumbling on the robot. This may be particularly the case, considering that collisions would occur within highly dynamic environments such as malls, airports, hospitals, markets, or mix-traffic areas where pedestrians, mobility devices, and even vehicles are frequent. Herewith, making a "frozen" robot a danger to itself and bystanders \cite{Salvini2021-soro}. In this work, we investigate a possible less dangerous and better accepted post-collision reaction that avoids the "freezing" robot problem. 

We extend our dynamical systems (DS) based obstacle avoidance controller \cite{Huber2019}, by augmenting it with a compliant control mechanism, using a passive DS approached offered in \cite{Kronander2016}. Assuming that we have real-time contact sensing in closed-loop control, we enable impedance control for our mobile robot through estimated contact forces over a known hull, following our previous approach for explicit force modulation for collaborative robot environments developed in \cite{Amanhoud2019}.
Unlike previous works on DS-based compliance, in our formulation the obstacle's exact shape and contact location are unknown. Thus, we simplify the problem by assuming a single contact point and use the only information at our disposal, namely the hull shape of the robot, for controlling the desired force during the interaction. 

A similar principle for closed-loop force control with 6-axes Force/Torque was offered in \cite{Kollmitz2018} with a focus on learning touch commands on a stiff hull. While the method in \cite{PaezGranados2017} offered an impedance control response when guiding a person's motion while limiting the maximum guidance force.
Our control approach offers the reactivity of time-invariant DS combined with contact estimations of impacts through a compliant bumper, herewith ensuring impact absorption through passive and active compliance for mitigating unexpected impacts with mobile robots.

We validate the method on the semi-autonomous standing mobility vehicle Qolo \cite{PaezGranados2018,Paez_tmech_2022} shown in Fig. \ref{fig:intro}; a type of powered wheelchair for standing mobility of lower-limb impaired people, similar to powered scooters, hoverboards, and unicycles, currently widespread.
We tested the approach to validate performance at mitigating contact forces by generating multiple collisions with a static obstacle varying the speed at contact. We show that the robot could successfully slide against the obstacle without exceeding the limit on contact force. 
Although current implementation relies on the assumption of single contact, this could be extended with higher sensing resolution such as multi-contact sensing in mobile manipulators through artificial skin \cite{Leboutet2019}, or through other sensing methods \cite{Kim2016}.
The source code and simulations with other robot types is available at: \href{https://github.com/epfl-lasa/sliding-ds-control}{\textit{https://github.com/epfl-lasa/sliding-ds-control}} \cite{compliant_ds_github}.

The remaining of this paper is organized as follows: We present the method and controller in section \ref{sec:problem}. We describe the controller structure with high-level obstacle avoidance in section \ref{sec:controller}. We evaluate the method in section \ref{sec:exp}.
Finally, we discuss and conclude in section \ref{sec:Discussion}.

\section{Problem Statement} \label{sec:problem}


In this formulation for post-collision control, we assumed: 
\begin{enumerate}
    \item Knowledge of the expected contact surface, namely a convex human body part.
    \item A collision could occur unexpectedly, thus, distance to the obstacle is unknown a priori.
    \item Expected contact occurs at a single location per sensing surface.
    \item The operational speed of the robot is slow enough to be safe in the transient phase thus, controllable post-collision.
\end{enumerate}
 
\begin{figure}[!t]
    \centering
    \includegraphics[width=7.5cm]{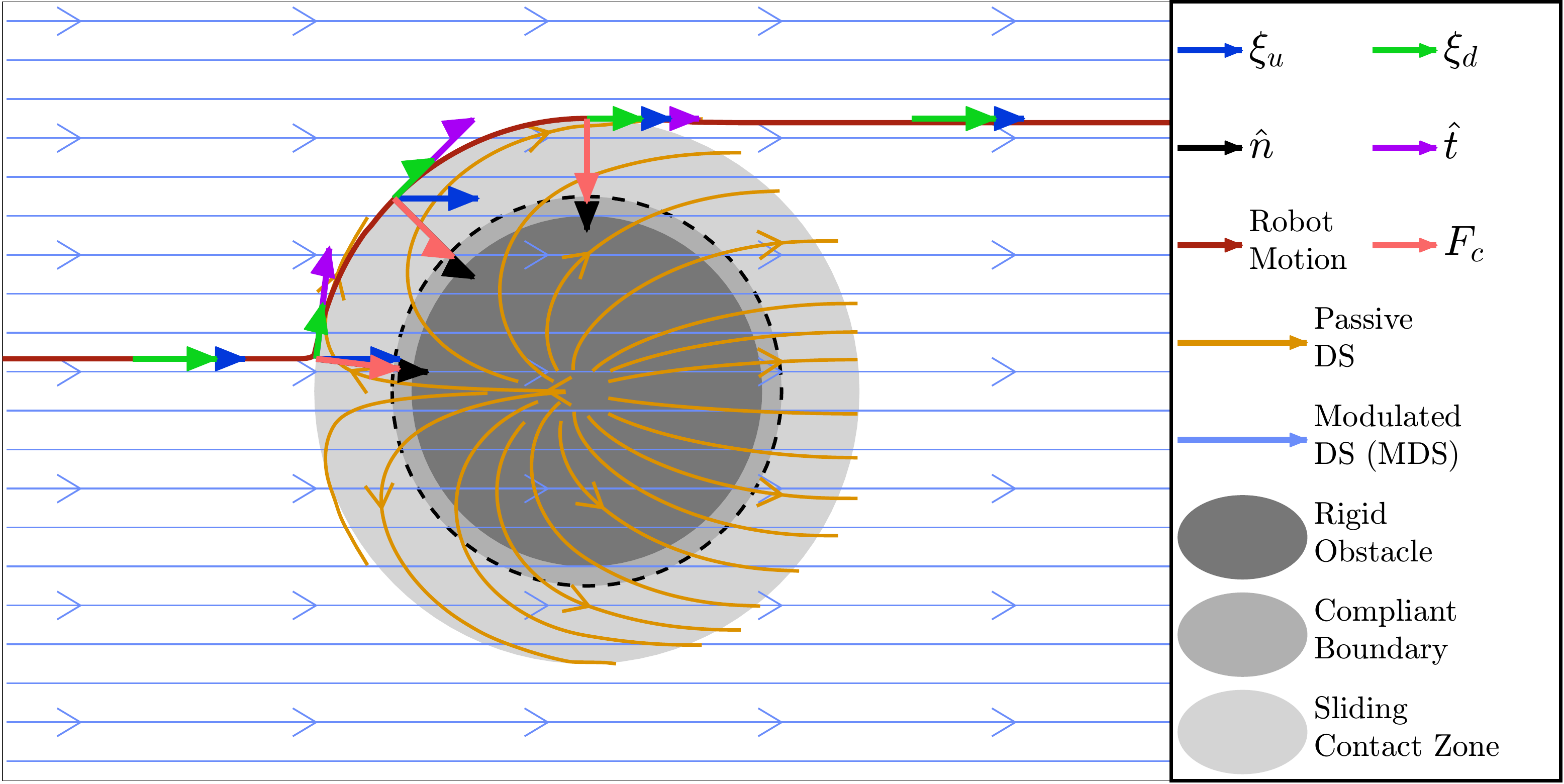}
    \caption{Sliding DS formulation for limiting contact forces while moving along an underlying desired motion. When the robot enters in contact with the obstacle (light-grey zone) the desired motion is controlled by the reaction force at the boundary guaranteeing a limited contact force $F_n$ to the obstacle and allowing a sliding motion $\xi_d$ around it. \label{fig:sDS}}
\end{figure}

In Fig. \ref{fig:sDS} we depicted a linear-DS with the robot represented as a holonomic point-mass (any point in this Cartesian space) and the pedestrian in contact as a convex shape. There are two zones of contact with the obstacle represented by: first, a physically impenetrable obstacle (dark grey), and second, a deformable region of the obstacle with a compliant boundary (dotted line) which allows controlling for safe contact force. 
Finally, we mark a sliding zone (lighter-grey) that represents the volume occupied by the robot during contact around the obstacle.
The resulting behaviour of the system is a force bounded sliding contact around the obstacle after entering in contact with the compliant boundary, assuming that there will be a state where the modulated DS will lead away from the contact surface without colliding with other obstacles.

\subsection*{Controller Formulation}

In order to derive the motion controller, we modelled the robot dynamics as: $M\ddot{\xi} + C\dot{\xi} = \tau_c + \tau_e$,
where $\dot{\xi} \in \mathbb{R}^2$ represents the robot's Cartesian velocity as a time invariant position dependent dynamical system.
$M \in \mathbb{R}^{2\times2}$ corresponds to the virtual mass of the robot, $C \in \mathbb{R}^{2\times2}$ accounts for centrifugal and Coriolis terms, $\tau_c$ represents the control forces and $\tau_e$ any external disturbances to be rejected.
An impedance-DS is used to achieve a sliding motion for overcoming the obstacle. We design a nominal trajectory controlling for the robot's speed and generated through a function $f({\xi}) \in  \mathbb{R}^{2}$. We then control for the torques using a damped tracking controller over  $f({\xi})$ ,
\begin{IEEEeqnarray}{rCl}
\tau_c &=&  \lambda_t f(\xi) -D \dot{\xi}
\end{IEEEeqnarray}
where $D \in \mathbb{R}^{2\times2}$ represents a negative defined damping effect.
$f(\xi)$ is the dynamical system effectively controlling the robot during contact, composed as: $f(\xi) = f_u(\xi) + f_n(\xi)$, 
where $f_u(\xi)$ represents the driving force generated by the nominal DS input tangential to the collision surface which can be transformed into the contact dynamics as:
    \begin{IEEEeqnarray}{rCl}
    f_u(\xi) &=& \hat{t}^T \frac{M \dot{\xi}_u}{T_s} \hat{t}
    \end{IEEEeqnarray}
where $T_s$ accounts for a discretizing time constant.
$f_n(\xi)$ describes the force control function as: 
    \begin{IEEEeqnarray}{rCl}
    f_n(\xi) &=& \frac{F_n + F_c}{\lambda_t} \hat{n}
    \end{IEEEeqnarray}
where $F_n $ was chosen a contact force limit bounded by safety and acceptability, while $F_c $ represents the measured contact force.
Yielding a controller of the form:
\begin{IEEEeqnarray}{rCl}
\tau_c &=& \lambda_t f_u(\dot{\xi}) + (F_n + F_c) \hat{n} - D \dot{\xi}
\end{IEEEeqnarray}

The damping effect on the matrix $D$ was controlled by a normal and tangential parameters over the surface of the obstacle, $\lambda_n$ and $\lambda_t$, respectively. 
\begin{IEEEeqnarray}{rCl}
D &=& Q \begin{bmatrix}
    \lambda_t & 0 \\
    0 & \lambda_n \\
\end{bmatrix} Q^T
\end{IEEEeqnarray}
where $ Q = \begin{bmatrix} \hat{t} & \hat{n}  \end{bmatrix}$ was defined by the contact location normal.
The parameter $\lambda_t$ allow to control the behaviour around the surface, for instance by setting $\lambda_t = 0$ we can provide an undamped free motion along the tangential direction of the collision surface.

Finally, transformation to the velocity domain of the robot was done by a first order Taylor expansion, thus the control for a desired velocity $\dot{\xi}_{d}$ can be written as,
\begin{IEEEeqnarray}{rCl}
\dot{\xi}_{d+1} &=& \frac{T_s}{M} \left(
    (F_n + F_c) \hat{n} - D \dot{\xi}_d
\right) + \hat{t}^T \dot{\xi}_u \hat{t} \label{eq:vel_cnt}
\end{IEEEeqnarray}
\subsubsection*{\textbf{Releasing contact:}}
Equation \ref{eq:vel_cnt} allows the robot to slide over the obstacle while limiting a constant contact force ($F_n$), but it does not allow the robot to move away from the obstacle. 
Thus, we defined an additional term in (\ref{eq:vel_cnt}) ($\hat{n}^T \dot{\xi}_u \hat{n}$) when the normal vector and underlying dynamical system desired motion oppose each other. 
Herewith, enabling the robot to move away from the obstacle if the DS indicates a feasible free-motion space.

Then, the final controller is defined as:
\begin{IEEEeqnarray}{rCl}
\dot{\xi}'_{d+1} &=&  \left\{
     \begin{array}{ll}
		\dot{\xi}_{d+1} + \hat{n}^T \dot{\xi}_u \hat{n} & \mbox{if } \left<\hat{n}, \dot{\xi}_u \right> < 0 \\
		\dot{\xi}_{d+1} & \mbox{otherwise}
	\end{array}
	\right.
\end{IEEEeqnarray}

\subsubsection*{\textbf{Velocity magnitude}}
When required by the robot application, the driving underlying DS magnitude could be upper bounded by $|\dot{\xi}_u| \ < |\dot{\xi}_{max}|$, hereby, guaranteeing a controllable speed allowed during contact interaction around the obstacle.

\subsubsection*{\textbf{Moving target}}
In case of a traceable moving obstacle in contact, we can include such estimation of the obstacle's motion to the previous formulation by defining the robot's state relative to the obstacle's pose ($\xi = x_r - x_o$). 
This effectively makes the desired motion $\dot{\xi_d}$ dependent on the obstacle's response. The DS, hence, speeds up or slows down according to the obstacle's speed while controlling the desired contact force.
Such behaviour requires explicit velocity estimation of the obstacle in contact. This estimate could be local, using, for instance, optical flow from a fish-eye camera or laser-based tracker of objects in close vicinity. While this requires the placement of more sensors, such as sensors scanning space on the side of the robot, this could enhance fluidity social navigation in interactions with a crowd flow or other dynamic obstacles.

\begin{figure}[!t]
    \centering
    \includegraphics[width=7.0cm]{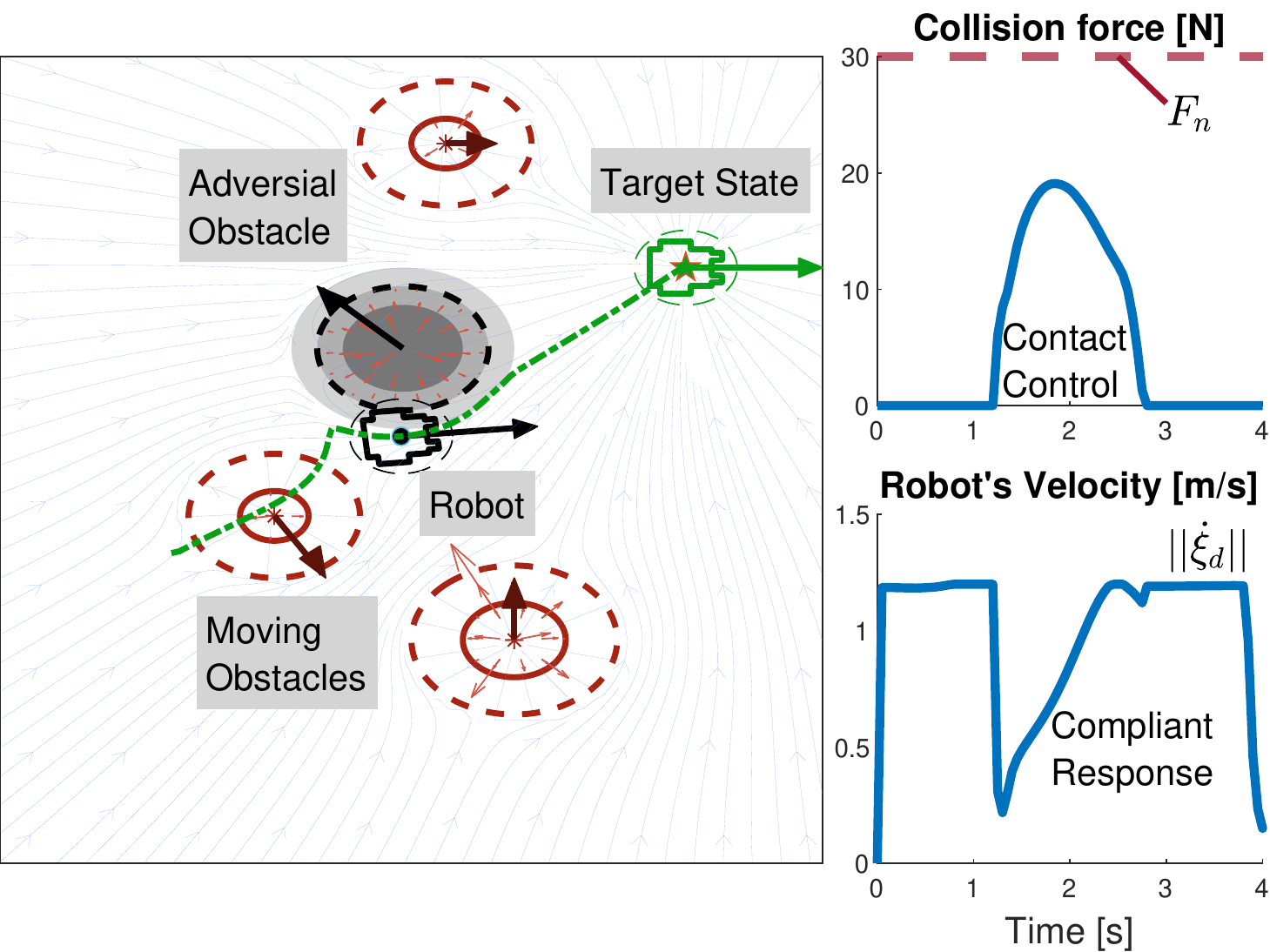}
    \caption{Sliding Dynamical System coupled with modulated obstacle avoidance example of a pre-collision and post-collision response. Here, multiple moving obstacles were modulated (red) in a linear-DS while a moving adversarial obstacle (grey) collided with the robot, forcing a sliding response while avoiding other obstacles.}
    \label{fig:MDS_SDS}
\end{figure}
Fig. \ref{fig:MDS_SDS} depicts a linear-DS towards an attractor (green mark) modulated by the surrounding moving obstacles. An example of a mobile robot navigation in 2D.
The resulting DS acts as the input to the proposed compliant modulation when an "adversarial" obstacle (invisible to the modulation) gets in contact (sensed by penetration and simulated with a constant mass-spring system), which triggers the compliant controller and enables a sliding behaviour around it while avoiding all other moving obstacles.

\section{Control architecture and Sliding surface }\label{sec:controller}

\begin{figure*}[!t]
    \centering
    \includegraphics[width=16cm]{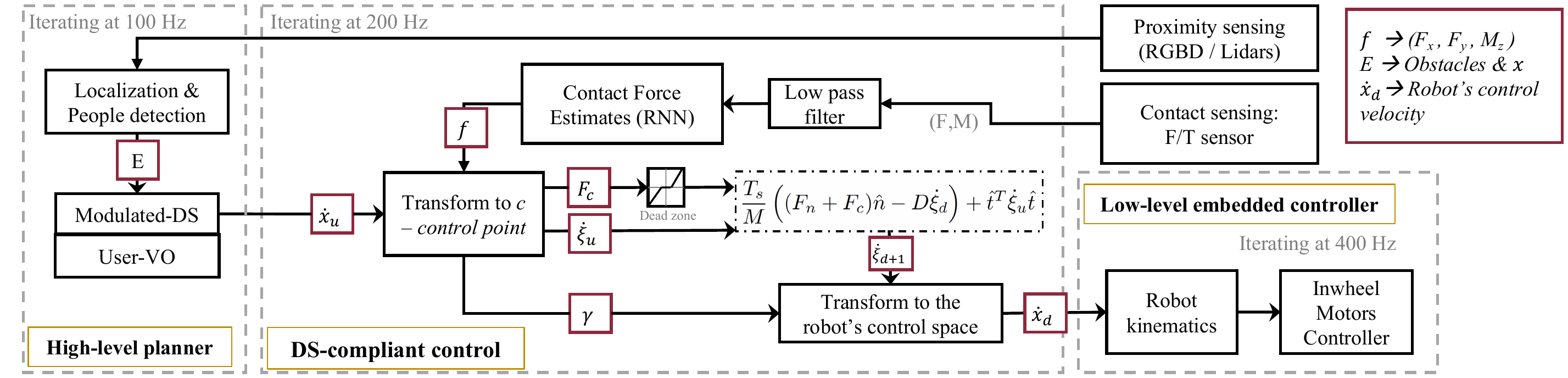}
    \caption{Compliant controller architecture proposed for the modulated and passive Dynamical System handling the post-collision through sliding.}\label{fig:controller}
\end{figure*}

We based the robot's control architecture on a continuous controller that handles three states of the mobile robot, namely: \textbf{obstacle avoidance}, \textbf{contact}, and \textbf{post-collision control}.
Fig. \ref{fig:controller} describes the general controller in three stages. First a high-level closed-loop motion planner drives the dynamics of the robot in its control space ($\dot{x}_u$).
While we used a modulated DS \cite{Huber2019} for obstacle avoidance, any other velocity-based planners could be used here, e.g, in the case of velocity obstacles (VO) based obstacle avoidance through the formulation in \cite{Gonon2021}. 
Second, control of compliance and contact force through sliding method using a known sensing surface over the robot's hull with a limited contact force $F_n$. Ensuring that the robot reacts to unexpected contacts and advances with a sliding manoeuvre should the underlying obstacle avoidance lead away from the contact surface without colliding with other obstacles.
Third, a low-level controller that handles the execution in real-time closed-loop control.

\subsection{Control Point over the Robot's Surface}
\begin{figure}[!t]
    \centering
	\includegraphics[width=5.5cm]{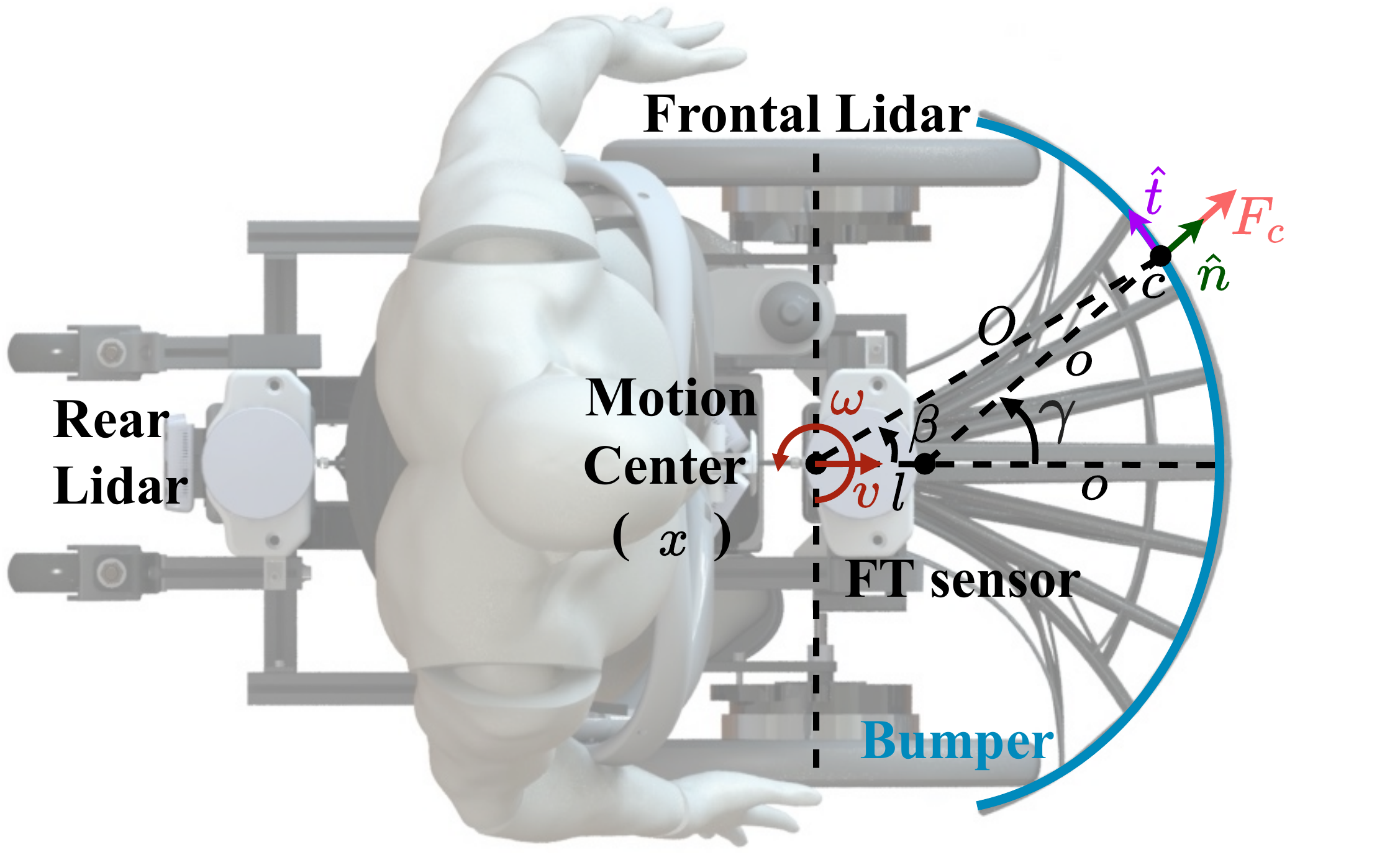}
	\caption{Implemented structure for compliant control in real-life scenarios through a frontal bumper on a person carrier type service robot. }\label{fig:qolo_bumper}
\end{figure}

To execute the control in a mobile robot we need to include a transformation between the robot's control space and a control point ($c$) around the bumper shape by using the estimated collision location over the bumper surface at an angle $\gamma$.
Which was estimated as $c = (o \cos{\gamma}, o \sin{\gamma})$ from the center of the bumper (see Fig. \ref{fig:qolo_bumper}). 

Thus, we control the effective velocity ($\dot{\xi}_{u}$) at the point of collision perpendicular to the bumper surface by transforming it to the control space of the robot.

The surface at the control point can be described using the normal ($\hat{n}$) and tangential ($\hat{t}$) unit vectors,
\begin{IEEEeqnarray}{rCl}
\hat{n} &=& (\cos{\gamma}, \sin{\gamma})
\\
\hat{t} &=& (-\sin{\gamma}, \cos{\gamma})
\end{IEEEeqnarray}

Then, we can define a Jacobian matrix ($J$) to transform the motion at center of the robot $x$ (in the case of Qolo, a non-holonomic differential wheel system) to an effective velocity ($\dot{\xi}_u$) at the control point on the surface in contact (holonomic):
\begin{IEEEeqnarray}{rCl}
\dot{\xi_u} &=& J \dot{x}_u
\\
&=& \begin{bmatrix}
    1 & -o \sin{\gamma} \\
    0 &  o \cos{\gamma}
\end{bmatrix} \begin{bmatrix} v_u \\ \omega_u \end{bmatrix}\nonumber
\end{IEEEeqnarray}

With a known hull surface (see Fig. \ref{fig:qolo_bumper}) described as,
\begin{IEEEeqnarray}{rCl}
    O &=& \sqrt{\left( o \sin{\gamma} \right)^2 + \left( l + o \cos{\gamma} \right)^2}
\\
    \beta &=& \tan^{-1}{\left( \frac{o \sin{\gamma}}{l + o \cos{\gamma}} \right)}
\end{IEEEeqnarray}
With the system controlled in the velocity space of the robot we map the DS through the following:
\begin{IEEEeqnarray}{rCl}
    \dot{\xi}_u &=& v_u \cos{\gamma} + \omega_u O \sin{\left( \gamma - \beta \right)}
\end{IEEEeqnarray}

where, $v_u$ and $\omega_u$ are the desired linear and angular velocities defined by the high level planner resp.

Finally, we transform back to the robot's control space ($\dot{x}_d = J^{-1} \dot{\xi}'_{d+1}$) through the inverse of the Jacobian.


\subsection{Non-linear compliance compensation} \label{subsec:bumper_compliance_comp}
The impact absorbing bumper (see Fig. \ref{fig:qolo_bumper}) was designed as a compliant surface light-weighted (\SI{900}{g}) in ABS material, mounted on a 6 axis force/torque (FT) sensor (Botasys Rokubi 2.0) at the centre of the semi-circular bumper.
The bumper was mounted through a padding material to the hard surface of the robot frame with spring-loaded screws at each side of the mount, thus, releasing part of the weight to the main frame of the robot. Herewith, mitigating the impact during the transient phase.
Unlike the work in \cite{Kollmitz2018} where a stiff hull was developed mounted on an FT sensor.
However, this brings a challenge in accurately estimating contacts on the surface of the hull, which we have tackled by learning the non-linearity of the passive-compliance through data of known impact forces.

To remove the effects of the non-rigidity of the bumper (see Fig. \ref{fig:qolo_bumper}), a prediction model was developed through 2 methods, first with a support vector regression (SVR), and secondly with a recurrent neural network (RNN), where we chose the second in implementation.

We trained the models with a known force applied to the bumper at various locations through an second FT sensor impacting the bumper surface. Then, the effective force and moment at the onboard FT sensor were estimated assuming a rigid model of the bumper.
The non-linear model was trained over the measured FT values from the onboard sensor to find $F_x$, $F_y$, and $M_z$.
We use three independent models to get corrections over $F_x$, $F_y$ and $M_z$.

\textbf{Assumptions:}
\textit{First}, other components of the forces ($M_x$, $M_y$ and $F_z$) are significantly smaller thus can be neglected.
\textit{Second}, the model removes all non-rigidity effects from the sensor measurements, herewith, we can consider the bumper as a rigid body on the robot's structure.
\textit{Third}, we only account for pure forces applied at the point of contact.

The calibration dataset consisted of slow and fast force variations on the bumper at various force magnitudes, impulse responses at various force magnitudes, and pulls on the bumper to cancel any undesired behaviour. 
The dataset was generated by applying a known force (using a FT sensor) on a $9 \times 4$ grid on the bumper. For each type of force profile, 3 samples of \SI{60}{s} of impacts were recorded at \SI{400}{Hz} at each point on the grid. Further details of the data and implementation of SVR and RNN are available in the online repository  \cite{compliant_ds_github}.



\paragraph*{\textbf{Contact force estimation comparison SVR vs. RNN}}
The calibration dataset is divided into $9:1$ split with $9$ part for training and $1$ part for validation. 
In this paper, on one hand, a $\nu$-SVR \cite{Kecman2005} was trained over the training dataset resulting in approximately $92,000$ support vectors for each axis. 
On the other hand, for the RNN, a single LSTM layer with a buffer of 6 samples was implemented. The output of the LSTM layer was then fed to a neural net for each of the desired force. Because of the lower estimation error ($< 15N$) and fastest processing time ($< 0.4 s$), we choose the RNN for real-time usage in the robot (as shown in table \ref{tab:estimation}).
\begin{table}[!t]
    \centering
    \renewcommand{\arraystretch}{1.3}
    \caption{Error Comparison between SVR and RNN for 3D Contact Estimation on the Compliant Bumper}
    \label{tab:estimation}
    \begin{tabular}{@{}llcc@{}}
        \toprule
          &  & SVR & RNN \\
        \midrule
        \multirow{3}*{Training Performance}
            & $F_x$ [\si{N}]   & $-0.81 \pm 15.82$ & $-0.19 \pm  8.25$   \\
            & $F_y$ [\si{N}]   & $21.68 \pm 42.53$ & $-0.95 \pm 13.08$   \\
            & $M_z$ [\si{N.m}] & $ 0.06 \pm 20.20$ & $-0.06 \pm  1.01$   \\
        \midrule
        \multirow{3}*{Testing Performance}
            & $F_x$ [\si{N}]    & $-3.93 \pm 31.67$ & $-0.16 \pm  9.99$   \\
            & $F_y$ [\si{N}]    & $36.55 \pm 46.55$ & $-4.36 \pm 14.99$   \\
            & $M_z$ [\si{N.m}]  & $-1.09 \pm 16.10$ & $ 0.10 \pm  0.82$   \\
        \midrule
        \multicolumn{2}{@{}l@{}}{On-board Computation Time [\si{\milli\second}]}
                           & $10.63$ & $0.383$          \\
        \bottomrule
    \end{tabular}
\end{table}

\subsection{Contact location estimation} \label{subsec:contant_loc_est}
We estimated the collision angle from the reference coordinate system at the sensor $\gamma$ from $F_x$, $F_y$, and $M_z$, as 
$M_z = F_x r \cos{\gamma} - F_y r \sin{\gamma}$.
Note that $F_x$ and $F_y$ at the point of contact and the sensor are assumed the same as described above. 
By replacing $\sin$ and $\cos$ by $\frac{e^{i\gamma} - e^{-i\gamma}}{2}$ and $\frac{e^{i\gamma} + e^{-i\gamma}}{2}$ respectively, we get eq. \ref{eqn:damper_transform_theta_cos_sin}. Eq. \ref{eqn:damper_transform_theta} describes explicitly the $\gamma$ by solving the quadratic equation in $e^{i\gamma}$.
\begin{IEEEeqnarray}{rCl}
\label{eqn:damper_transform_theta_cos_sin}
    \frac{M_z}{r} &=& 
    F_x \left( \frac{e^{i\gamma} + e^{-i\gamma}}{2} \right) +
    i F_y \left( \frac{e^{i\gamma} - e^{-i\gamma}}{2} \right)
\\
    \label{eqn:damper_transform_theta}
    \gamma &=& -i \log{\left(\frac
        {\frac{M_z}{r} + i\sqrt{F_x^2 + F_y^2 - \left( \frac{M_z}{r} \right)^2}}
        {F_x + i F_y}
    \right)}
\end{IEEEeqnarray}

Finally, the force magnitude $F_{mag}$ can be estimated as $F_{mag} = F_x \sin{\gamma} + F_y \cos{\gamma}$.

\section{Experimental Evaluation} \label{sec:exp}

For evaluation in combination with obstacle avoidance in close to real-life expected situations, we equipped the robot Qolo \cite{Paez_tmech_2022} with 2 Lidars (Velodyne VLP-16), and an RGBD  sensor (Intel Realsense). Obstacle's tracking was performed through real-time people detection implemented through a pipeline of sensing fusion of Lidar-based detection by DR-SPAAM \cite{Jia2020} and RGBD detection through YOLO \cite{yolo_2021}.
The full controller repository can be found here: \href{https://github.com/DrDiegoPaez/qolo_ros.git}{\textit{Qolo-ROS}} \cite{Paez_Qolo_github}.
The control parameters were set as follows, he discretizing time constant ($T_s$) to the sampling time of the control loop (\SI{200}{Hz}) for real-time controller dynamics. The virtual robot mass ($M$) was set to \SI{2}{kg} for light-behaviour. Damping parameters $\lambda_t$ and $\lambda_n$ were set to $0$ and $0.5$ respectively to allow undamped tangential motion and partially damped normal motion of the collision surface.

\begin{figure}[!t]
    \centering
    \subfigure[Collision force.  \label{fig:forces}]{\includegraphics[width=4.2cm]{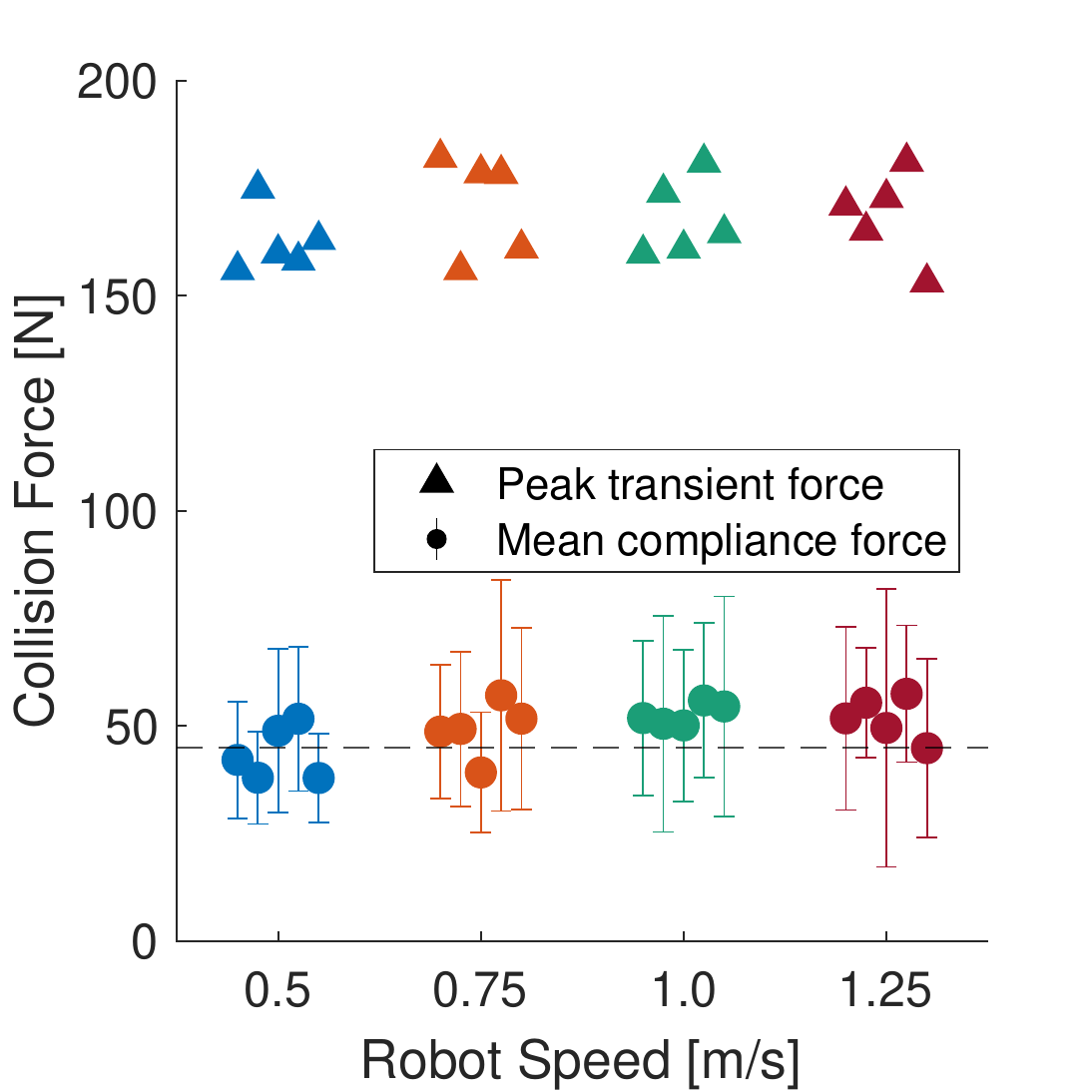}}
	\hfill
    \subfigure[Time until stable compliant control (transient time).\label{fig:times}]{\includegraphics[width=4.2cm]{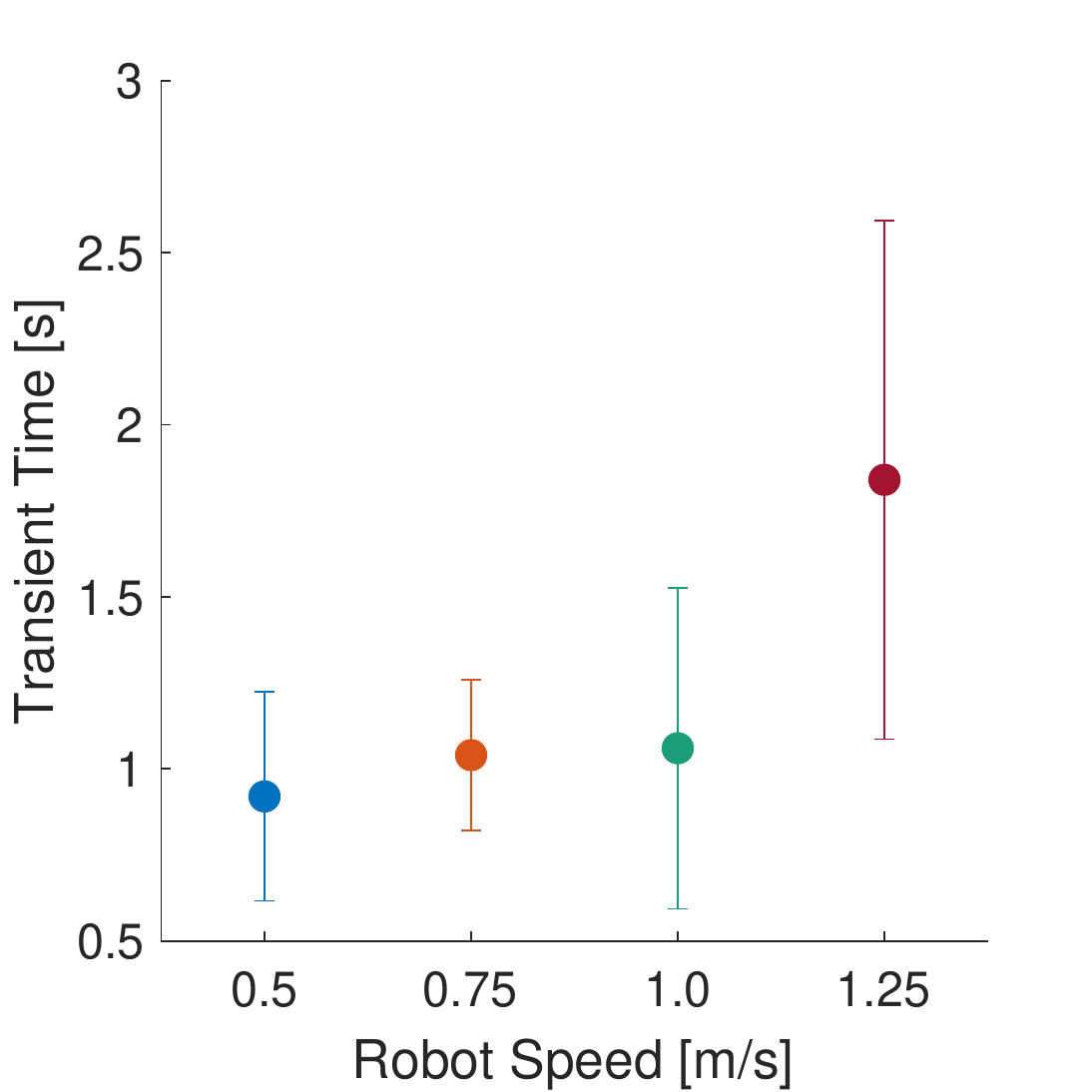}} 
    \caption{Resulting post-collision contact force response to changes in the operational speed at the time for achieving a stable active compliance.}
    \label{fig:collision_multiple_speed}
\end{figure}

\subsection{Contact Control Evaluation at Multiple Speeds}
First, we evaluated the effects of increasing the operational speed of the robot on the post-collision force response with the proposed controller, to understand the effect of the approach for real-life contact situations.
We run five sets of collisions per condition between the mobile robot Qolo (\SI{45}{kg}) and a static person (\SI{80}{kg}).
The contact force limit for compliance was set to $F_n = \SI{45}{N}$ (well below the average pain threshold for the lower legs \SI{130}{N} \cite{ISO15066}). While the desired motion ($\dot{\xi}_u$) was set to a linear-DS (ignoring the obstacle) towards an attractor at \SI{4.5}{m} ahead of the robot.

The tests were conducted at four operational speeds of \SIlist{0.5;0.75;1.0;1.25}{m/s}, by setting the robot to approach the person (only the authors) frontally and about \SI{0.1}{m} of the line of motion, thus impacting the bumper at $\gamma = \pi / 6$. 
The post-contact speed magnitude $|\dot{\xi}_{max}|$ was set to \SI{0.5}{m/s}.

We observed a peak collision force over \SI{150}{N} in most cases, as expected for the transient phase (see, Fig. \ref{fig:forces}), given the overall delay of the control system (passive-compliance is needed for absorbing part of the impact).
The mean collision force was \SI{52 \pm 9}{N} for speeds up to \SI{1}{m/s}, and \SI{51 \pm 15}{N} for the highest speed of \SI{1.25}{m/s}. All contact forces during sliding were well within the desired limit of \SI{65}{N} ($50\%$ of the pain threshold).
Results are summarized in table \ref{tab:results}.
The mean transient phase time (time to stabilise the compliance control) showed a difference between impacts below \SI{1}{m/s} with \SI{1.06 \pm 0.46}{s}, while for impacts at \SI{1.25}{m/s} presented a slower response of \SI{1.84 \pm 0.75}{s} (see, Fig. \ref{fig:times}). 
Further control parameter settings are detailed in the supplementary repository \cite{compliant_ds_github}.
\subsection{Evaluation with Integrated Obstacle Avoidance}
\begin{figure*}[!t]
     \centering
         \subfigure[Snapshot of the experiment where an adversarial pedestrian intentionally collides with the robot while a driver sets the desired velocity. ]{\includegraphics[width=17.0cm]{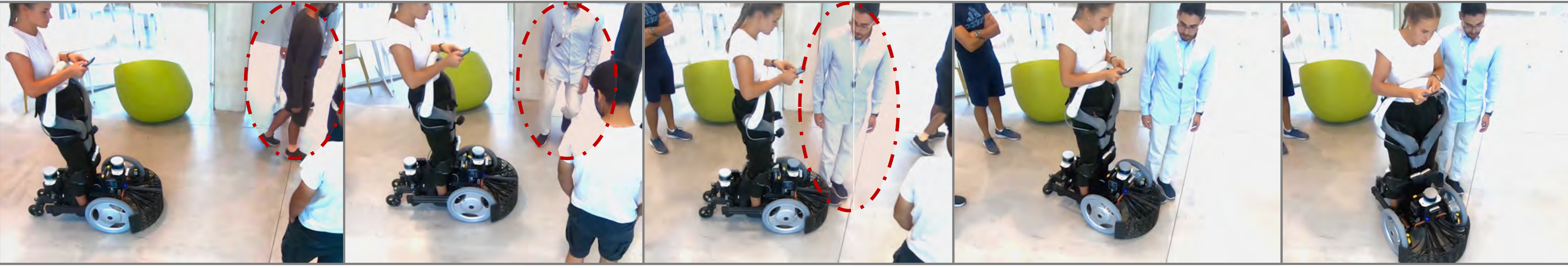}%
    		\label{fig:photos}}
		\hfil 
		\subfigure[Velocity obstacles based control with a user a high-level input, a limiting contact force $F_n=\SI{30}{N}$, resulting in \SI{25.88 \pm 8.90}{\newton}.]{\includegraphics[width=8.5cm]{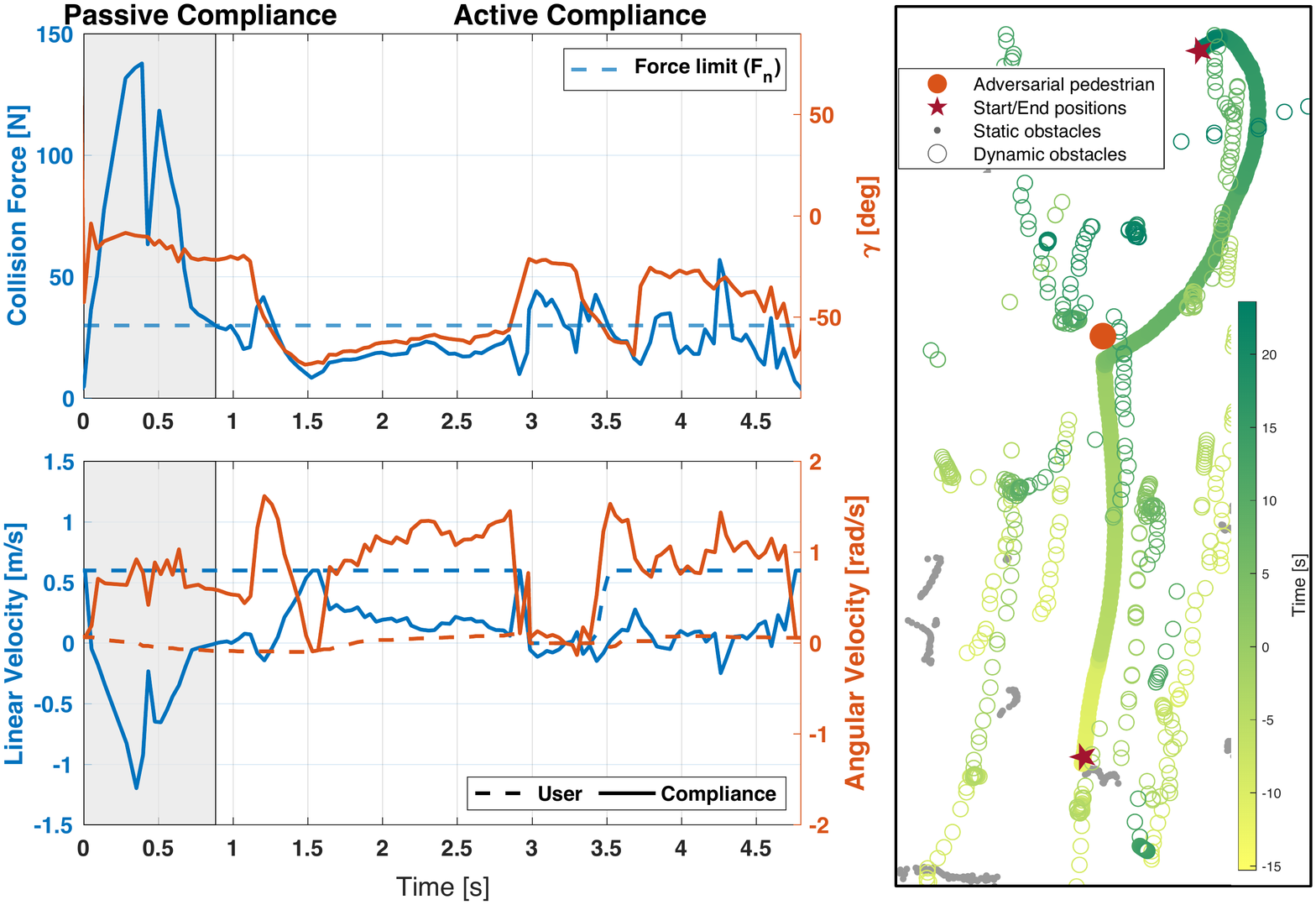}%
		\label{fig:vo}}
		\hfil 
    	\subfigure[Modulated DS mode with a limiting contact force $F_n=\SI{45}{N}$, resulting in \SI{49.18 \pm 11.97}{\newton}. ]{\includegraphics[width=8.5cm]{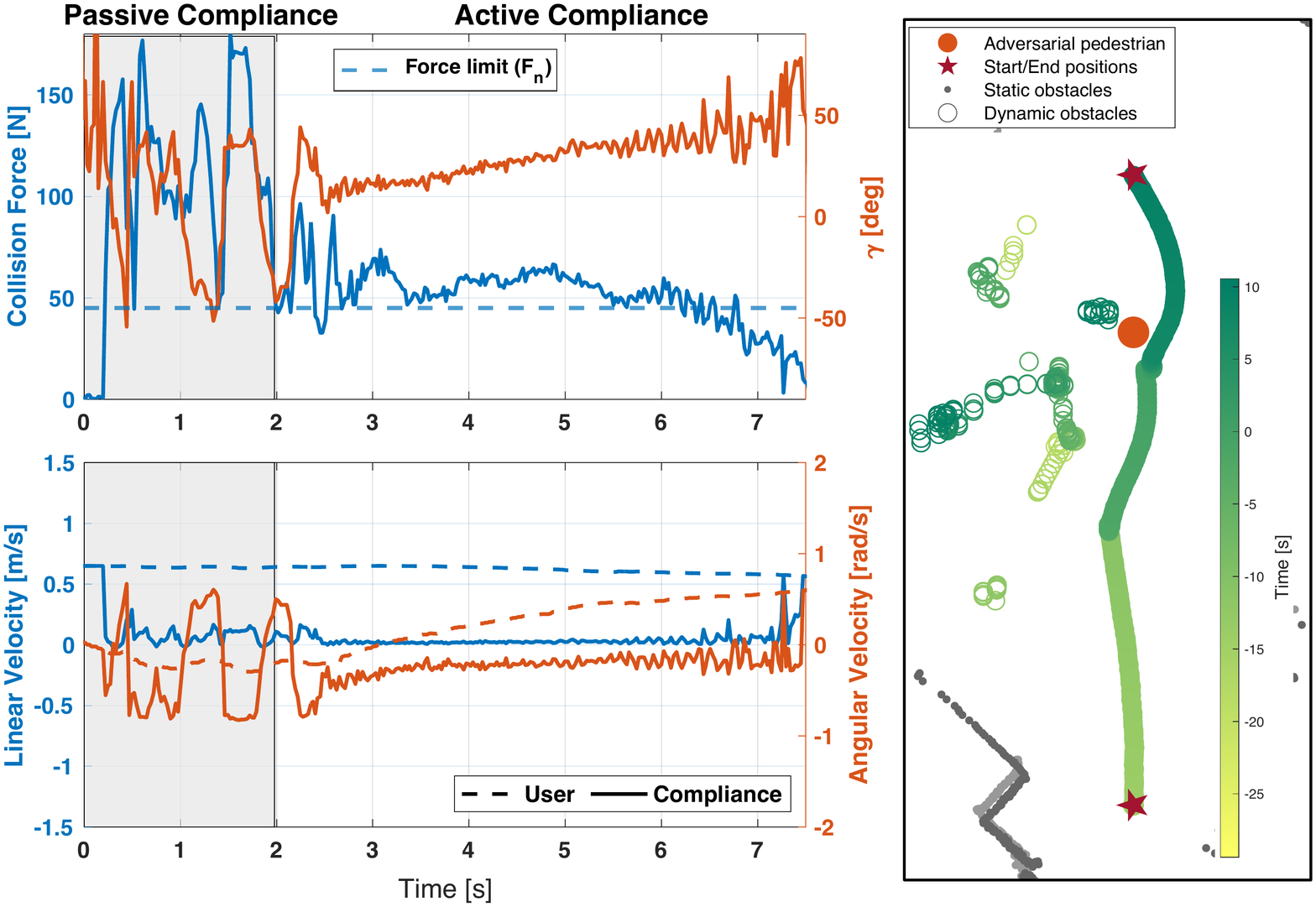}%
		\label{fig:mds}}
		\hfil
		\caption{Experimental setup of obstacle avoidance and post-collision control scenario with an adversarial pedestrian. }\label{fig:exp_sample}
\end{figure*}

We further tested the approach in autonomous driving mode and contrasted two obstacle avoidance methods: first modulated obstacles DS from \cite{Huber2019} and second, a variant on the Velocity-obstacle (VO) that solves operator given commands with the known shape of the robot \cite{Gonon2021}.
We recorded three tests per method with a safe operational velocity ($\dot{\xi}_u = \SI{0.65}{m/s}$) and sliding velocity ($|\dot{\xi}_{max}|=\SI{0.5}{m/s}$). 
Results in Fig. \ref{fig:photos} shows an illustration of the experimental scenario, where an adversarial pedestrian jumps in front of the robot from behind another pedestrian faster than the physical limits of the robot's actuation, thus producing a collision.
In both examples (see Fig. \ref{fig:vo}-\ref{fig:mds}) there is the first stage of transient impact (left figure, shaded area) beyond any control on the robot (passive-compliance is needed for absorbing part of the impact). 
Subsequently, the compliant controller limits the contact interaction (active compliance phase) to the set force value ($F_n$), while performing the sliding motion around the obstacle as long as there is no other pedestrian on its path.
\begin{table}[!t]
    \centering
    \caption{Post-collision force statistics} 
    \label{tab:results}
    \begin{tabular}{@{} c S[table-format=1.2] S[table-format=2.2] S[table-format=2.2]@{}}
        \toprule
            & &
            \multicolumn{1}{c}{Mean Force [\si{N}]} & 
            \multicolumn{1}{c}{Std. Deviation [\si{N}]}  \\
        \midrule
            \multirowcell{4}{Robot\\ Speed\\\.[\si{m/s}]}
                & 0.5   & 42.74   & 6.18    \\
                & 0.75  & 48.04   & 9.01    \\
                & 1.0   & 52.51   & 9.07    \\
                & 1.25  & 51.64   & 15.37   \\
        \midrule
            \multirowcell{2}{Control\\ Type}
                & \multicolumn{1}{c}{VO control} & 46.08 & 6.97 \\
                & \multicolumn{1}{c}{MDS Control} & 54.57 & 5.63 \\
        \bottomrule
    \end{tabular}
\end{table}
Results showed performance of contact force error of \SI{9.5 \pm 5.6}{N}, and  \SI{1.0 \pm 6.9}{N} for MDS and VO modes resp. (see, table \ref{tab:results}). In VO mode (with an operator on-board) the heavier robot was more controllable, achieving a smaller error in comparison with MDS mode.
Results of the trajectory response showed in VO mode that the user was able to overcome the adversarial pedestrian by drifting away quicker from contact, thus trajectories deviated further from the linear path (see, Fig. \ref{fig:vo} right side).
In contrast, the MDS control made longer contact with the obstacle, moving around it, thus the overall trajectories were closer to the original linear DS (see, Fig. \ref{fig:mds} right side).

\section{Summary and Discussion} 
\label{sec:Discussion}
We have presented a control method for a mobile service robot to achieve a reactive control on post-collision which allows to absorb part of the impact and continue moving by sliding around the pedestrian. Herewith, proposing an alternative solution to the common "safe" approach of freezing a robot upon contact. 

The experimental assessment at multiple operational speeds allowed us to illustrate the approach in real-life operations. In the case of the robot Qolo, we found the current control system capable of recovering from an unexpected collision within \SI{1}{s} of the impact when driving below \SI{1}{m/s}, and navigating around the unforeseen obstacle.

The results with integrated obstacle avoidance showed a robot able to overcome a pedestrian in both tested methods, namely, modulated dynamical systems (DS) and velocity obstacles (VO). In both cases without incurring further contact with bystanders, and within desired limits of "safe" contact force.
Among the two methods, a smoother trajectory was achieved in DS controlled motions, whereas user-driven VO tests showed lower contact forces. A likely result of the user decision of drifting away quicker from contact, which in turn resulted in a larger deviation from the initial trajectory.

Future work should look more closely at how to handle multiple contacts. As well, investigate how pedestrian responses to contact and its effect to the robot planner for social acceptability and safety.

\section*{Acknowledgement}
This work was funded by the EU H2020 project "Crowdbot" (779942).
The experiments were approved with an ethical protocol by the human research ethical committee of EPFL (Approval No: HREC-032-2019).
We thank K. Suzuki and the University of Tsukuba for lending the robot Qolo used in this experiment. \textit{Disclaimer:} DP holds a patent of the robot Qolo and shares in the company Qolo Inc.


%

\bibliographystyle{IEEEtran}
\bibliography{IEEEabrv,Compliant_control_post_collision-ICRA2021}

\end{document}